\documentclass[conference]{IEEEtran}
\IEEEoverridecommandlockouts

\usepackage{amsfonts}
\usepackage{amsmath}
\usepackage{amssymb}
\usepackage{flushend} 
\usepackage{booktabs}
\usepackage{cite}
\usepackage{graphicx}
\usepackage[letterpaper,margin=0.75in,columnsep=0.2in]{geometry}
\usepackage{fancyhdr}
\usepackage{placeins} 
\usepackage{tabularx}
\usepackage{times}
\usepackage{xcolor}
\usepackage[hidelinks]{hyperref}

\AtBeginDocument{%
  \setlength{\textfloatsep}{14pt plus 2pt minus 2pt}%
  \setlength{\floatsep}{12pt plus 2pt minus 2pt}%
  \setlength{\intextsep}{12pt plus 2pt minus 2pt}%
  \setlength{\dbltextfloatsep}{14pt plus 2pt minus 2pt}%
  \setlength{\dblfloatsep}{12pt plus 2pt minus 2pt}%
  \setlength{\abovecaptionskip}{6pt plus 1pt minus 1pt}%
  \setlength{\belowcaptionskip}{0pt}%
}

\makeatletter
\renewcommand{\section}{\@startsection{section}{1}{\z@}%
  {3.0ex plus 1.2ex minus 0.2ex}%
  {1.25ex plus 0.3ex}%
  {\normalfont\normalsize\centering\scshape}}
\renewcommand{\subsection}{\@startsection{subsection}{2}{\z@}%
  {2.0ex plus 0.8ex minus 0.2ex}%
  {0.95ex plus 0.25ex}%
  {\normalfont\normalsize\itshape}}
\renewcommand{\subsubsection}{\@startsection{subsubsection}{3}{\z@}%
  {1.5ex plus 0.5ex minus 0.1ex}%
  {0.7ex plus 0.15ex}%
  {\normalfont\normalsize\itshape}}
\makeatother

\newcommand{\act}{\boldsymbol{a}}
\newcommand{\ind}{\mathbf{1}}
\newcommand{\Acal}{\mathcal{A}}
\newcommand{\Mcal}{\mathcal{M}}
\newcommand{\Ccal}{\mathcal{C}}
\title{\texorpdfstring{\vspace*{0.25in}}{}\Large\bfseries Action-Conditioned VLM Monitoring for Autonomous Vehicle Semantic Anomalies}

\author{\IEEEauthorblockN{Daniel Adejumo\textsuperscript{2}, Omkar Ankalkope\textsuperscript{2}, Kunal Runwal\textsuperscript{2}, and Aliasghar Arab\textsuperscript{1,2,$*$}}
\thanks{\textsuperscript{1}Aliasghar Arab is with The City College of New York, Grove School of Engineering, New York, NY 10031, USA. aarab@ccny.cuny.edu}%
\thanks{\textsuperscript{2}Daniel Adejumo, Omkar Ankalkope, Kunal Runwal, and Aliasghar Arab are with the Department of Mechanical and Aerospace Engineering, Tandon School of Engineering, New York University, 6 MetroTech Center, Brooklyn, NY 11201, USA. aliasghar.arab@nyu.edu}%
\thanks{\textsuperscript{$*$}Corresponding author.}%
}

\begin{document}
  \maketitle
  \thispagestyle{fancy}

  \begin{abstract}
    We study \emph{action-conditioned monitoring} of autonomous-vehicle
    behavior in semantically unusual driving scenes. The target is whether
    the executed ego action is appropriate for the scene: the same scene
    can admit both normal and anomalous behavior, so visual unusualness
    alone does not determine the monitoring label. A Vision-Language
    Model (VLM) first predicts the scene-required action from an
    outcome-hidden video window; the monitor then compares that
    expectation with the realized ego-motion trajectory under an explicit
    compatibility rule. To study this monitoring task, we contribute a
    six-field taxonomy of 112 paired scenarios spanning both failure
    polarities (42 derived from documented incidents), a generated video
    corpus with estimated trajectories, and an exploratory
    two-stage monitor on that corpus. On 235 clips from 15 scenario
    families, the two-stage protocol with a fixed $T-2.5$\,s window
    obtains 83.0\% accuracy and anomaly recall 0.773, versus 74.5\% and
    0.536 for a single-stage video-only baseline; specificity is 0.880
    versus 0.928.
  \end{abstract}


  \section{Introduction}
  \label{sec:intro}
  Autonomous Vehicles (AVs) can fail even when a familiar object is
  perceived correctly: a stop sign is printed on a billboard, brake lights
  appear on a video display, or a charged fire hose resembles flat debris.
  These \emph{semantic anomalies}~\cite{elhafsi2023semantic} arise when an
  object appears outside its nominal context and the resulting ego
  response is unsafe or unnecessary. Documented fire-hose, flooded-road,
  and school-bus incidents show that such contextual failures have
  practical safety consequences~\cite{npr2023driverless,nhtsa2026waymorecall,ntsb2026schoolbus}. Foundation models have been used broadly in robotic autonomy stacks, from converting agent states into text with a Large Language Model (LLM) to limit planned speed~\cite{wang2025dualad} to reasoning over semantic scene descriptions for anomaly detection in a reactive planner~\cite{elhafsi2023semantic}.


  Pixel- and object-level benchmarks detect unfamiliar content or unusual
  traffic events~\cite{blum2021fishyscapes,chan2021segmentmeifyoucan,li2022coda,yao2023dota},
  but do not directly evaluate whether a familiar object's displaced
  context makes the executed ego action appropriate. AnoVox broadens
  simulated anomaly evaluation to multimodal content and temporal
  anomalies and includes ego action states~\cite{bogdoll2024anovox};
  large-scale AV corpora provide egomotion and action-grounded reasoning
  labels~\cite{nvidia2026physicalai}, and retrieval-augmented Vision Language Models (VLM)
  support corner-case comprehension~\cite{wang2026rac3}. Operational Design Domain (ODD) based scenario trees are generated with LLMs and then instantiate in CARLA simulations to evaluate both OOD recognition and
  safe-action selection in~\cite{aasi2025generating}. Thus, reasoning about
  an appropriate action in an unusual scene is already addressed. Our
  distinction is to assign \emph{Normal/Anomaly} labels to \emph{executed}
  scene-action pairs, using an outcome-hidden expectation and separately
  recovered ego motion. Each scenario specification supplies both appropriate and inappropriate responses.

  Scene understanding has been used for VLM-based scene descriptions which is connected to a regulation
  database to select lawful future
  paths~\cite{han2025traffic}. Their rule checks guide candidate-plan
  selection. Our explicit rule instead compares a scene-required
  expectation with a canonical class of realized motion to produce a
  behavioral verdict. It is a coarse action-compatibility relation, not
  a jurisdiction-specific traffic-law compliance test.

  A two-phase, layered analysis of real driving
  images is studied in~\cite{brusnicki2025savant}. Related driving-oriented
  models map scenes toward future actions using
  VLMs~\cite{xu2024drivegpt4,tian2024drivevlm,nvidia2025cosmosreason,bhat2026vlmautodrive}.
  These works do not evaluate whether the realized ego response matches
  an expectation inferred before the label-defining outcome. In the
  context-displacement cases of interest~\cite{elhafsi2023semantic}, the
  relevant object remains visually familiar but its role changes with
  context. This differs from pixel-level open-set
  segmentation~\cite{blum2021fishyscapes,chan2021segmentmeifyoucan} and
  lies within the ISO~21448 space of unsafe behavior without a component
  fault~\cite{iso21448}.

    \begin{figure}[t!]
    \centering
    \includegraphics[width=.9\columnwidth]{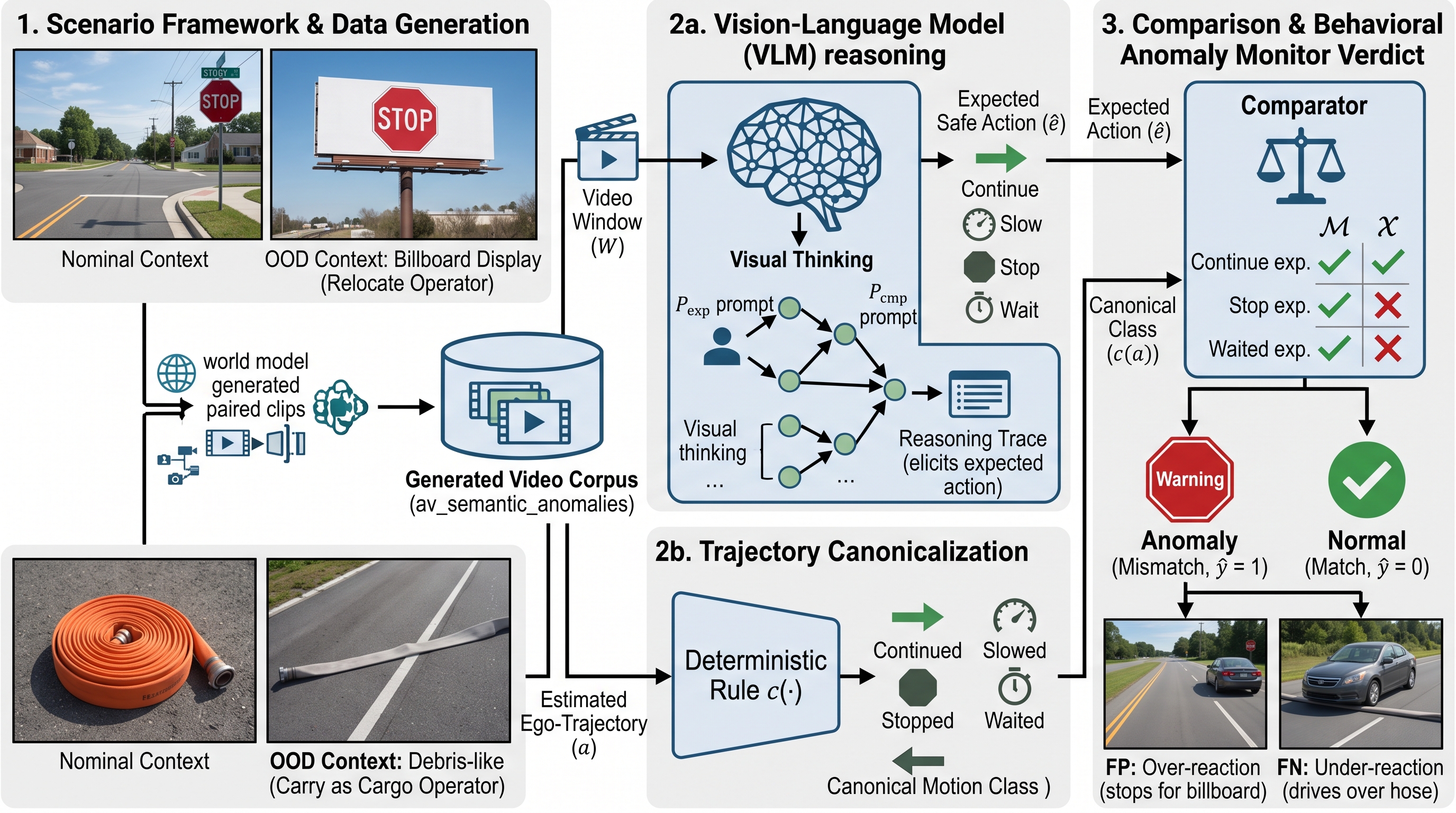}
    \caption{Action-conditioned monitoring. Context displacement defines
      paired scene-action labels; a VLM infers the scene-required action
      from an outcome-hidden window, then compares it with canonicalized
      ego motion.}
    \label{fig:framework}
  \end{figure}
  
In this work, we formulate the task as \emph{action-conditioned monitoring}: given a driving scene and realized ego motion, assess whether the vehicle's response matches what the scene requires. Inspired by scenario-based testing~\cite{menzel2018scenarios}, continuing past a depicted stop sign can be appropriate, whereas braking for that depiction can
  be anomalous; a charged fire hose across the lane reverses the required
  response. The monitoring label therefore belongs to a scene-action
  pair. Fig.~\ref{fig:framework} outlines the monitor: it infers the
  scene-required action from an outcome-hidden video window, then checks
  compatibility with the observed ego motion. This separates scene interpretation from the
  assessment of the vehicle's response.
  
  The main contribution of this research is an action-conditioned monitoring formulation
  in which the verdict expresses compatibility between a scene-required
  response and realized ego motion. Other contributions include: (i) a two-stage VLM
  protocol that separates scene-based expectation inference from explicit
  action-compatibility checking; (ii) a six-field taxonomy of 112 paired
  scenarios, including 42 derived from documented incidents, with both
  failure polarities attached to each row; and (iii) a generated ego-video
  corpus with ordinal action labels from a prompt heuristic. An optical-flow
  probe and human review support exploratory evaluation of
  monitor accuracy.
  
  \section{Problem formulation and Method}
  In this work, a \emph{semantic behavioral anomaly} is defined as a mismatch between
  the action required by a scene and the ego action that is executed. The monitor assigns a behavioral verdict to a scene-action pair by
  checking whether the realized motion is compatible with the response inferred from the scene. Let $V$ be an ego-view clip of duration $T$ and
  $\act=((v_1,\psi_1),\ldots,(v_K,\psi_K))$ its 5-Hz estimated
  trajectory tuple, where $v_i$ is estimated ego-vehicle speed and $\psi_i$ is the estimated ego-vehicle heading angle at the $i$th trajectory sample. Let $\Acal=\{\text{Continue},\text{Slow},\text{Stop},
  \text{Wait}\}$ and $\Ccal=\{\text{continued},\text{slowed},
  \text{stopped},\text{waited},\text{moved off}\}$. The
  context-aware action is $e^*\in\Acal$, stage 1 predicts $\hat e\in\Acal$, and the canonical motion class is~$c(\act)\in\Ccal$. Hence, the evaluation
  label $y$ is the intended polarity of the generated clip. It is not recomputed from recovered motion. The monitor prediction $\hat y$ is defined as
  \begin{equation}
    \hat y = \ind[c(\act)\notin\Mcal(\hat e)],\label{eq:label}
  \end{equation}
  where the reported protocol uses
  \begin{equation}
    \begin{aligned}
      \:\:\:\:&\Mcal(\mathrm{Continue})=\Mcal(\mathrm{Slow})=\{\mathrm{continued},\mathrm{slowed}\},\\
      &\Mcal(\mathrm{Stop})=\{\mathrm{stopped},\mathrm{waited}\},\\
      &\Mcal(\mathrm{Wait})=\{\mathrm{waited}\}.
    \end{aligned}\label{eq:compat}
  \end{equation}
  Because Continue and Slow are mutually compatible,
  applying~\eqref{eq:compat} to $e^*$ would label a Continue-versus-Slow
  contrast as Normal even when the taxonomy tag is Anomalous. Both stages run zero-shot learning~\cite{qwen2026qwen38},
  selected from eight configurations,
  using vLLM~\cite{kwon2023efficient} on large computational units. Each clip uses a separate conversation, a fixed
  seed, and large computational budgets. Video is sampled at
  $8$~fps and Lanczos-upscaled from $832\times480$ to $1248\times720$.

  \subsubsection{Stage 1: expectation from an outcome-hidden window}
  A limited-horizon window of the video is defined as
  \begin{equation}
    W = V\big[\max(0,\; t^* - \Delta),\; t^*\big],\label{eq:window}
  \end{equation}
  where $\Delta = 2.5~\mathrm{s}$ is a constant window duration and $t^*$ is the end time of the window. The primary protocol
  ends the window at a fixed cutoff $t^*=T-2.5$~s
  in~\eqref{eq:window}, before the label-defining outcome. Thus explicit
  trajectory input and the final outcome are withheld, although
  pre-cutoff visual motion may remain visible. The prompt also includes
  a depiction guard directing the model to distinguish physical hazards
  from depictions. Annotated per-clip decision times are used only in the offline oracle ablation.
  
  \subsubsection{Stage 2: action comparison}
  The second turn supplies $P_{\mathrm{cmp}}$ and a deterministic
  narrative $g(\act)$ containing the canonical motion class, endpoint
  speeds, and, when applicable, heading change. The prompt asks the VLM to apply
  the compatibility rule in~\eqref{eq:compat}. Given $\hat e$ and
  $c(\act)$, the verdict is fully determined by Eq.~\eqref{eq:label}.
  Narrative rendering removes a numeric-reading step that is difficult
  for language
  models~\cite{gruver2023large,spathis2024first}.

The prompt $P_{\mathrm{exp}}$ represents the Stage 1 prompt used to elicit the expected action ($\hat{e}$), mirroring the $P_{\mathrm{cmp}}$ prompt used in Stage 2. Both stages emit a delimited reasoning trace~\cite{wei2022chain}, but only the final answer is scored. Missing or truncated decisions count as errors without resampling; none are reported in the two-stage 235-clip runs. 

   \begin{table}[t!]
    \caption{The six-field semantic-anomaly framework}
    \centering
    \footnotesize
    \begin{tabularx}{\columnwidth}{@{}c l X@{}}
      \toprule
      \# & Field & Description \\
      \midrule
      1 & Object in Driving Scene & Entity with a mandated response (stop sign, brake lights, pedestrian, fire hose) \\
      2 & Nominal Context & Where the object normally lives; reacting to it there is correct \\
      3 & OOD Context & Object or look-alike displaced outside its nominal context, appearance preserved \\
      4 & Normal Action in OOD & Context-aware response: react to the scene, not the pattern \\
      5 & Anomalous Action in OOD & Failure mode: react to the pattern as if nominal, or miss a real hazard presenting atypically \\
      6 & Example Scenario & One-sentence description; the generation / evaluation prompt seed \\
      \bottomrule
    \end{tabularx}
    \label{tab:framework}
  \end{table}

      \begin{table}[h!]
    \caption{Example scenario pairs, one per failure polarity}
    \centering
    \footnotesize
    \setlength{\tabcolsep}{3pt}
    \renewcommand{\arraystretch}{1.08}
    \begin{tabularx}{\columnwidth}{@{}>{\raggedright\arraybackslash}p{0.6in} >{\raggedright\arraybackslash}X >{\raggedright\arraybackslash}X@{}}
      \toprule
      Field & Fire hose & Brake lights \\
      \midrule
      Nominal context & Coiled on a fire engine, off the roadway & Rear of a lead vehicle in the ego lane \\
      OOD context & Charged hose across an active lane at a fire scene & Life-size car ad with glowing brake lamps on a night digital billboard \\
      Normal action & Stop before the hose; never cross; wait to be waved through & Recognize the screen as media; hold speed and lane \\
      Anomalous action & Classifies it as debris; drives over the hose & Reads a braking lead vehicle; brake-checks at speed \\
      Example & A car drives over a charged fire hose at an active fire scene & A car brake-checks for brake lights on a nighttime video billboard \\
      Tag & REAL $\cdot$ FN & SYN $\cdot$ FP \\
      \bottomrule
    \end{tabularx}
    \label{tab:examples}
  \end{table}
  \begin{figure*}[ht!]
    \centering
    \includegraphics[width=\textwidth]{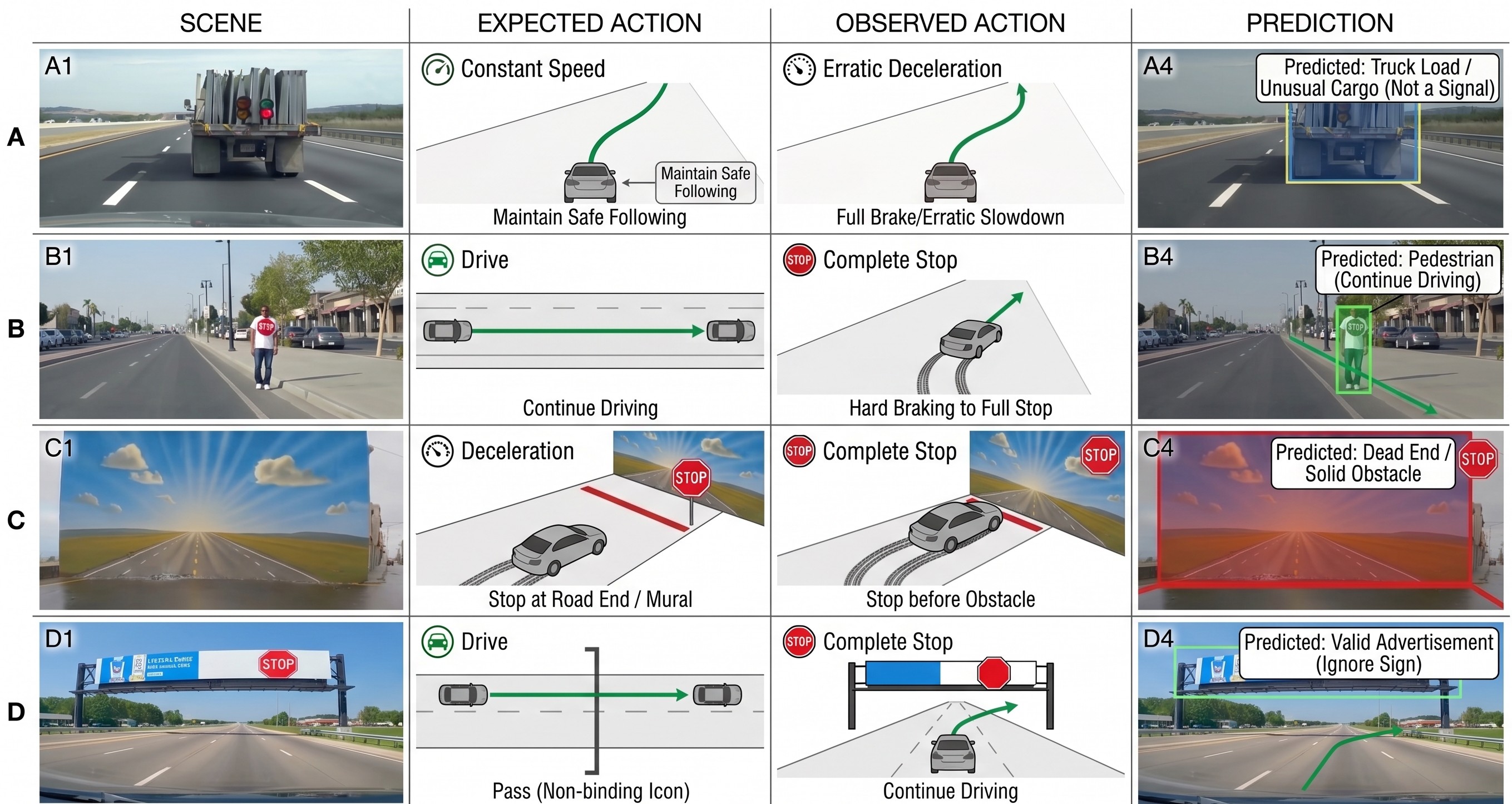}
    \caption{Illustrative synthetic driving scenarios: (A) traffic signals
      carried as cargo, (B) a stop-symbol depiction on clothing, (C) a
      road mural, and (D) a stop-symbol advertisement. The columns
      juxtapose the scene, expected action, depicted ego response, and
      scene-level interpretation.}
    \label{fig:synthetic_examples}
  \end{figure*}
  \subsection{Scenario-based Framework}
  \label{sec:method}
  The proposed scenario-based testing framework parametrizes each case around an entity, its context, and the resulting behavioral contrast. Every scenario decomposes into the six fields of
  Table~\ref{tab:framework}, each field's output feeding the next.  Fields 1--3 build the scene: a semantically loaded object with a
  mandated response, its nominal context, and a displacement of the
  object or a look-alike into an OOD context,
  appearance preserved, via reusable operators. Fields 4--5 give the behavioral contrast: the Normal Action of a
  context-aware driver versus the Anomalous Action of a pattern-matching
  stack. Field 6 collapses the row into a one-sentence scenario used as the generation seed. Because fields 4--5 attach both behaviors to one scenario
  specification, every row defines a Normal/Anomalous pair. Generated clips are \emph{scenario-matched}; we do not assume they are pixel-identical counterfactuals. The taxonomy covers \emph{false-positive} (FP) failures that act on a non-threat and \emph{false-negative} (FN) failures that ignore an atypically presented hazard. Table~\ref{tab:examples} shows one row of each polarity.

  Each row carries a provenance tag. REAL denotes derivation from a
  documented incident, not real imagery. Examples include robotaxis
  interfering with active fire scenes~\cite{npr2023driverless}, a
  flooded-roadway failure associated with a 3{,}791-vehicle
  recall~\cite{nhtsa2026waymorecall}, and an automated vehicle passing an
  active school bus stop signal~\cite{ntsb2026schoolbus}. SYNTHETIC rows
  extrapolate the same mechanisms to new object-context combinations.
  The framework yields a 12-row seed taxonomy and a 100-row extension:
  112 pairs across ten categories, with 42 REAL and 70 SYNTHETIC rows. The categories cover traffic-control look-alikes; pedestrian,
  humanoid, and animal confusions; weather and lighting; road surfaces
  and markings; unusual cargo and vehicles; emergency scenes; adversarial
  projections; reflections and displays; and infrastructure or
  operational anomalies. The tabular form supports provenance tracking,
  polarity balancing, and generation from each row's Example field.

  \subsection{Synthetic Video Generation}
  \label{sec:videogen}
  Semantic anomalies are rare, and deliberately collecting balanced
  unsafe behavior is impractical. We therefore synthesize the corpus
  with the Cosmos~3 world foundation model~\cite{nvidia2026cosmos3}.
  Each taxonomy row is rendered separately with its Normal and Anomalous
  action descriptions. An LLM expands each seed into a structured prompt;
  two rounds of human review check realism while preserving the intended
  anomaly. Scenery variants change weather, lighting, and setting.
  Cosmos3-Nano generates 24-fps, $832\times480$ clips, mostly 5~s long
  (21 are 8~s), with a label-defining action segment of at least 2~s.
  Reviewers check the object, context, action, and visual artifacts;
  failed clips are regenerated or excluded. Fig.~\ref{fig:synthetic_examples} illustrates four context-displacement
  scenarios and their scene-action contrasts.

  \label{sec:action}
  The Cosmos~3 video-to-action inverse-dynamics pipeline used for estimating trajectories, executed with
  Cosmos3-Nano, estimates speed and heading from each generated clip.
  Its 5-Hz inference sequence supports ordinal motion classification;
  pipeline speed units are not calibrated vehicle telemetry. Clip
  admission requires agreement between a prompt-level action heuristic,
  a non-contradictory optical-flow probe, and human review of the
  synchronized video and trajectory. Failed clips are regenerated or
  excluded; 68 clips entered the release through this gate.

  \label{sec:vlm}
  \begin{figure*}[h!]
    \centering
    \includegraphics[width=\textwidth]{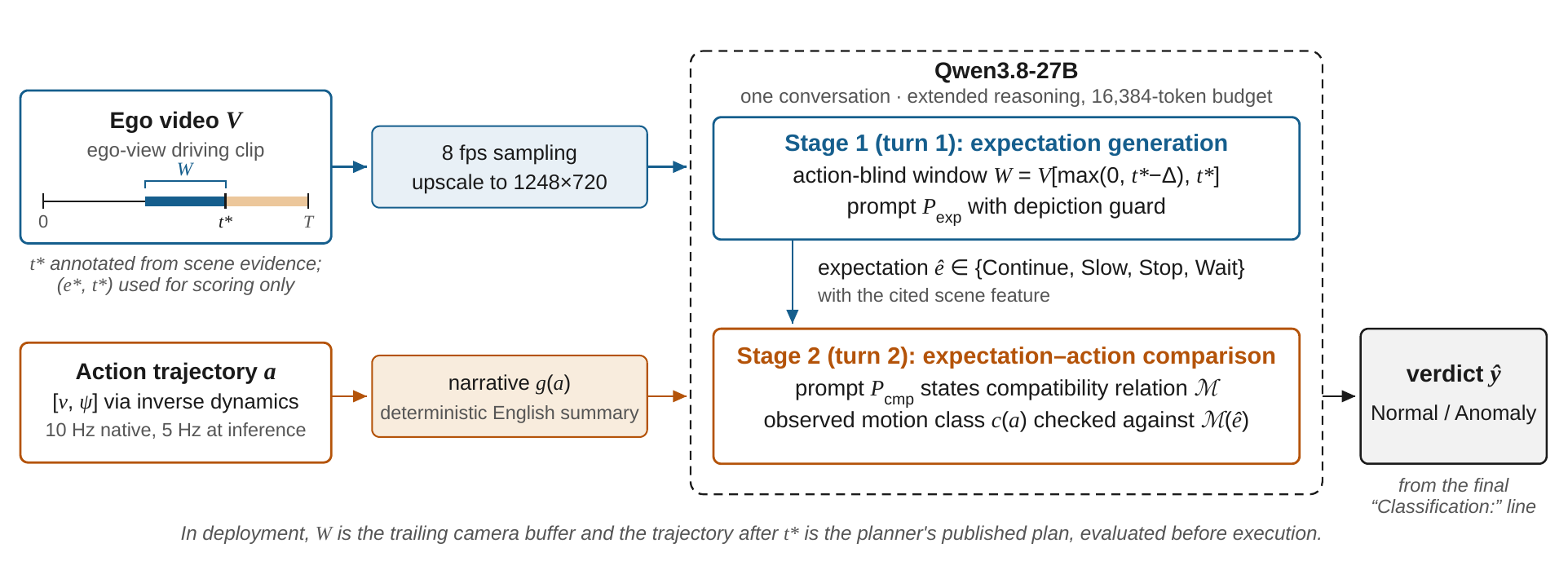}
    \caption{Action-conditioned monitoring through expectation-action
      comparison. Stage 1 predicts the
      scene-required action from an outcome-hidden window; stage 2
      compares the canonicalized ego motion with an explicit
      compatibility relation.}
    \label{fig:pipeline}
  \end{figure*}

  \section{Validations and Results}
  \label{sec:results}
  \label{sec:dataset}
  The \textit{av\_semantic\_anomalies} corpus contains 315 generated
  ego-view videos and 630 trajectory files (1.09~GB), organized as
  scenery-varied families from the seed taxonomy. The primary comparison uses 235 clips (110 Anomalous, 125 Normal) from
  15 scenario families; individual ablations use the stated 72- or
  80-clip subsets. Admission is the three-signal ordinal check of
  Section~\ref{sec:action}. The two-stage monitor of
  Fig.~\ref{fig:pipeline} is compared with a single-stage video-only
  baseline that uses
  the same model and decoding budget, with no explicit trajectory,
  compatibility rule, or two-stage factorization. The results demonstrate balanced accuracy, anomaly recall, normal specificity, and
  accuracy with clip-level 95\% Wilson intervals. Paired comparisons use
  the two-sided exact McNemar test where discordances are available.

  \begin{table*}[t]
    \caption{Exploratory monitoring results on 235 clips
      (110 Anomalous, 125 Normal): primary comparison and offline oracle ablation.}
    \centering
    \footnotesize
    \begin{tabularx}{\textwidth}{@{}Xccccc@{}}
      \toprule
      Method & \shortstack{Balanced\\accuracy $\uparrow$}
      & \shortstack{Anomaly\\recall $\uparrow$}
      & \shortstack{Normal\\specificity $\uparrow$}
      & \shortstack{Accuracy\\(correct/total)}
      & \shortstack{Accuracy\\95\% Wilson CI} \\
      \midrule
      Single-stage video-only (baseline)
      & 0.732 & 0.536 (59/110) & \textbf{0.928} (116/125)
      & 0.745 (175/235) & [0.685, 0.796] \\
      Two-stage, fixed $T-2.5$~s (ours, primary)
      & \textbf{0.826} & \textbf{0.773} (85/110) & 0.880 (110/125)
      & \textbf{0.830} (195/235) & [0.777, 0.872] \\
      Primary $-$ baseline (pp)
      & $+9.4$ & $+23.6$ & $-4.8$ & $+8.5$ & -- \\
      \midrule
      Two-stage, annotated $t^*$ (offline oracle)
      & 0.818 & 0.764 (84/110) & 0.872 (109/125)
      & 0.821 (193/235) & [0.767, 0.865] \\
      Annotated $t^*$ $-$ baseline (pp)
      & $+8.6$ & $+22.7$ & $-5.6$ & $+7.7$ & -- \\
      \midrule
      Majority class (reference)
      & 0.500 & 0.000 (0/110) & 1.000 (125/125)
      & 0.532 (125/235) & -- \\
      \bottomrule
    \end{tabularx}
    \par\vspace{2pt}
    \begin{minipage}{\textwidth}\footnotesize
      \emph{Notes:} Balanced accuracy averages anomaly recall and
      specificity. Bold marks the better primary-comparison value;
      $\uparrow$ means higher is better. Differences are percentage
      points (pp), computed before rounding from class counts derived
      from the reported rates and totals. Accuracy intervals are
      clip-level 95\% Wilson intervals without scenario-family
      clustering adjustment.
    \end{minipage}
    \label{tab:monitor}
  \end{table*}
  
  Table~\ref{tab:monitor} foregrounds balanced accuracy, anomaly recall,
  and normal specificity. For the fixed $T-2.5$~s window, the primary
  protocol achieves 0.826 balanced accuracy versus 0.732 for the
  video-only baseline, a gain of 9.4 percentage points. Anomaly recall
  increases by 23.6 points (0.536 to 0.773), while specificity decreases
  by 4.8 points (0.928 to 0.880). These changes are computed before
  rounding the rates. Accuracy improves from 175/235 to 195/235.
  Replacing that cutoff with an annotated
  per-clip decision time $t^*$ scores 193/235. The two prediction sets
  differ on 25 clips; paired discordances for that comparison are not
  provided. Given an accepted
  expectation, the Stage-2 comparator is correct on 180/181 clips
  (99.4\%): once $\hat e$ and $c(\act)$ are known, the VLM almost always
  applies~\eqref{eq:compat}. Stage-wise attribution still requires a
  joint error table.

  Against the baseline, paired discordances are available only for the
  annotated-$t^*$ variant: 44 baseline errors corrected and 26
  baseline-correct cases degraded (net $+18$, exact McNemar $p=0.041$).
  No paired discordances are available for the primary fixed-window run,
  so no $p$-value is attached to its improvement.
  Fig.~\ref{fig:example} gives one example.

  \begin{figure}[b]
    \centering
    \fbox{\begin{minipage}{\dimexpr\columnwidth-2\fboxsep-2\fboxrule\relax}\footnotesize
    Stage 1 (early window only): \emph{``Looking at the scene, the red
    octagonal `STOP' symbol is painted on an overhead billboard
    [\dots], not an actual traffic sign. A genuine stop sign would never
    be mounted on a billboard gantry spanning a free-flowing highway.
    Since it is only an image, it commands no action. [\dots] a correct,
    safe driver should simply maintain current speed and lane.
    Continue''}

    \smallskip
    Stage 2 (trajectory narrative $g(\act)$ = ``the vehicle slowed from
    about 45 mph to a complete stop and was stationary at the end.''):
    \emph{``Observed motion: the vehicle decelerated from highway speed
    to a full stop and remained stationary at the end (stopped). Since I
    expected the driver to continue at speed (the `STOP' sign was just a
    painted billboard image, so nothing real required braking), the
    actual stop is a mismatch. Classification: Anomaly, the behaviour
    does not match the expected action''}
    \end{minipage}}
    \caption{Worked billboard example. Selected output excerpts are
      reproduced verbatim; omissions are marked by ellipses and
      reasoning traces are omitted.}
    \label{fig:example}
  \end{figure}

  Missed anomalies decrease from 51 to 25, while false alarms increase
  from 9 to 15. Total errors decrease from 60 to 40. The protocol detects 65/70 FP-style anomalies (spurious response to a non-threat) and 19/40 FN-style anomalies (failure to react to an
  atypical hazard) on the annotated-$t^*$ variant, giving recalls of
  $0.929$ and $0.475$. Of that variant's 42 errors, 21 come from one mural
  scenario family, for which the video-only baseline is stronger ($19/31$
  vs.\ $10/31$). Sixteen errors involve Stop/Slow or low-speed-creep
  granularity, and five are isolated cases. We do not have the
  corresponding polarity split for the 40 errors of the primary
  fixed-window run. On an 80-clip class-balanced development subset, three decoding seeds
  give $68/80$, $68/80$, and $64/80$ for the two-stage protocol versus $57/80$,
  $56/80$, and $57/80$ for the baseline. Mean accuracy is
  $83.3\%\pm2.9$ points versus $70.8\%\pm0.7$ points (sample standard
  deviation across seeds), with a mean paired gain of $12.5\pm3.3$
  points. We do not pool the repeated predictions into an artificial
  $n=240$ independent-sample test. Median end-to-end latency on the annotated-$t^*$ two-stage run is
  46.9~s (90th percentile $110.6 s$) on the H200. With concurrency 6, the
  235 clips complete in $39.5 min$, equivalent to $10.1~s$ of wall-clock
  throughput per completed clip. The single-stage baseline has $61.8 s$
  median latency.

  \label{sec:ablations}
  The following ablations use their stated 72-, 80-, or 235-clip
  subsets; results from different subsets are not directly comparable.
  Paired predictions are unavailable for the action-form and
  token-budget comparisons, as are fold-level paired predictions for
  fine-tuning; no significance claims are made for these comparisons. Table~\ref{tab:models} compares eight configurations on 72 clips.
  Qwen3.8-27B gives the highest early-window strict expectation count,
  31/72, versus 18/72 for Cosmos3-Nano (a 13-clip or 18.1-point
  increase). Scaling from Cosmos3-Nano to Super does not improve this
  count. The full-clip minus early-window difference $\Delta$ is used as
  an outcome-anchoring diagnostic: Qwen checkpoints vary by $-2$ to
  $+3$ clips, whereas InternVL3.5 increases by 17. This diagnostic is
  suggestive rather than a calibrated anchoring measure.

  \begin{table}[h!]
    \caption{Model comparison on the 72-clip selection subset: stage-1
      expectation (early window), outcome-anchoring diagnostic, and
      single-stage verdict}
    \centering
    \footnotesize
    \setlength{\tabcolsep}{2.5pt}
    \renewcommand{\arraystretch}{1.05}
    \begin{tabularx}{\columnwidth}{@{}Xrrrr@{}}
      \toprule
      Model & Early S/L & Full S & $\Delta$ & Verdict acc. \\
      \midrule
      Cosmos3 (Nano, driving-tuned)~\cite{nvidia2025cosmosreason,nvidia2026cosmos3} & 18 / 38 & 22 & +4 & 0.736 \\
      Cosmos3 (Super, driving-tuned) & 8 / 26 & 12 & +4 & -- \\
      Qwen3.8-27B (selected) & 31 / 46 & 30 & $-1$ & 0.597 \\
      Qwen3.6-27B & 30 / 41 & 28 & $-2$ & 0.694 \\
      Qwen3.5-27B & 29 / 42 & 32 & +3 & 0.639 \\
      Qwen3-VL-32B-Thinking & 19 / 37 & 20 & +1 & 0.681$^{\mathrm{a}}$ \\
      GLM-4.6V-Flash & 11 / 28 & 13 & +2 & 0.528 \\
      InternVL3.5-38B & 22 / 34 & 39 & +17 & 0.653 \\
      \bottomrule
    \end{tabularx}
    \par\vspace{2pt}
    \begin{minipage}{\columnwidth}\footnotesize
      $^{\mathrm{a}}$Reasoning-mode verdict; all other verdict entries
      use direct response.
    \end{minipage}
    \label{tab:models}
  \end{table}

  Table~\ref{tab:preproc} varies frame rate and representation size on
  72 clips. At native resolution, $4$ to $8$~fps changes accuracy from 0.639
  to $0.611$ ($-2.8$ points), providing no observed benefit. At the
  matched 8~fps rate, upscaling changes accuracy from $0.611$ to $0.736$
  ($+12.5$ points). The 0.639-to-0.736 comparison ($+9.7$ points)
  changes both factors and is not an isolated upscaling effect.

  \begin{table}[h!]
    \caption{Input preprocessing (single-stage video-only, 72-clip subset)}
    \centering
    \footnotesize
    \begin{tabular}{@{}lcc@{}}
      \toprule
      Input & Visual prompt tokens & Acc. \\
      \midrule
      $832\times480$, 4~fps & ${\approx}\,4.0$k & 0.639 \\
      $832\times480$, 8~fps & ${\approx}\,8.0$k & 0.611 \\
      $1248\times720\,{\uparrow}$, 8~fps & ${\approx}\,11.7$k & 0.736 \\
      \bottomrule
    \end{tabular}
    \label{tab:preproc}
  \end{table}

  Table~\ref{tab:action} varies the Stage-2 action representation on
  $80$ clips. Raw $[[v,\psi]]$ with free-form comparison gives $49/80$;
  adding $\Mcal$ gives 54/80; velocity-only input gives 53/80; and
  narrative plus $\Mcal$ gives $67/80$. Thus the rule adds five net
  correct predictions, removing heading changes the net count by $-1$,
  and the narrative adds $13$ relative to raw input with the same rule.
  Appending raw trajectories to a one-stage prompt reduces the reported score,
  consistent with a numeric-reading
  difficulty~\cite{gruver2023large,spathis2024first}, but does not by
  itself establish why individual predictions change.

  \begin{table}[h!]
    \caption{Stage-2 action representation (two-stage, 80-clip
      development subset; stage 1 fixed)}
    \centering
    \footnotesize
    \setlength{\tabcolsep}{4pt}
    \renewcommand{\arraystretch}{1.08}
    \begin{tabularx}{\columnwidth}{@{}>{\raggedright\arraybackslash}Xcc@{}}
      \toprule
      Stage-2 formulation & Acc. & Comparator$^{\mathrm{a}}$ \\
      \midrule
      Raw $[[v, \psi]]$, free-form comparison & 0.613 & ${\approx}\,0.70$ \\
      Raw $[[v, \psi]]$ + explicit rule $\Mcal$ & 0.675 & ${\approx}\,0.83$ \\
      Raw $[v]$ only + rule $\Mcal$ & 0.662 & ${\approx}\,0.83$ \\
      Narrative $g(\act)$ + rule $\Mcal$ (ours) & 0.838 & 63/63 = 1.00 \\
      Video only (no action; necessarily single-stage) & 0.750 & -- \\
      \bottomrule
    \end{tabularx}
    \par\vspace{2pt}
    \begin{minipage}{\columnwidth}\footnotesize
      $^{\mathrm{a}}$Accuracy conditional on an accepted expectation.
    \end{minipage}
    \label{tab:action}
  \end{table}

  On the full $235$ clips, the guard changes direct accuracy from $0.562$
  to $0.621$ and reasoning-mode accuracy from 0.621 to 0.745. Because it
  was selected among more than 30 tested wordings on this set, these
  post-hoc gains require confirmation on held-out scenarios.

  Reasoning mode is associated with gains of $12.4$ points with the guard
  and 5.9 points without it. At $8{,}192$ tokens, approximately $10\%$ of
  outputs reportedly truncate; scoring only completed outputs gives
  0.815 and is biased. Counting truncations as errors, the reported
  72-clip score changes from $52/72$ ($0.722$) to $58/72$ (0.806) when the
  budget increases to $16{,}384$. On the reported subset, shortening the stage-1 window from $2.5$ to $1.5$
  and 1.0~s changes accuracy from 0.613 to 0.600 and 0.588. Extending
  the window toward the outcome recovers 12--14 of the 42
  annotated-$t^*$ errors but exposes label-defining motion, so it is not
  a fair outcome-hidden condition. Majority-of-three stage-1 sampling gives $66/80$ versus $68/80$
  for one sample. Rank-16 Low-Rank Adaptation (LoRA) stage-1 tuning with five-fold
  leave-scene-group-out cross-validation gives aggregate final
  accuracies of $0.736$ and $0.728$ for answer- and procedure-style
  targets, below the annotated-$t^*$ zero-shot 0.821. A within-scenario split
  gives 0.936, indicating that random variant splits substantially
  overestimate transfer.

  \FloatBarrier

  \subsection{Discussion and Limitations}
  \label{sec:limitations}
All evaluated footage is generated,
  including incident-derived REAL scenarios. The results do not measure
  transfer to real driving, and recovered motion supports ordinal
  classes rather than calibrated vehicle telemetry. The protocol
  evaluates realized motion offline. Automatic triggering,
  planner-trajectory monitoring with a trailing camera
  buffer~\cite{sinha2024realtime}, and closed-loop intervention remain
  unevaluated; the measured latency does not establish real-time
  feasibility. Independently generated polarity
  clips are matched by scenario specification, but scene appearance and
  ego motion may covary. A video-only baseline can exploit either source
  of information. The full-protocol comparison also changes trajectory
  input, the compatibility rule, and reasoning structure together.
  Without a trajectory-swap or controlled same-scene action intervention,
  the observed gains do not isolate the causal contribution of action
  conditioning. The action-form ablations vary representation, not the
  association between scene and action. Family clustering and prompt selection on the
  235-clip set limit statistical inference: clip-level intervals and
  $p$-values are descriptive. Leave-scene-group-out fine-tuning measures
  supervised transfer within the available families; it does not provide
  a held-out test of the zero-shot monitor. Confirmation requires a
  locked scenario-disjoint test set.

  \section{Conclusion}
  \label{sec:conclusion}
  We formulated action-conditioned monitoring as an explicit comparison
  between a scene-required expectation and realized ego motion. A six-field taxonomy of paired semantic-anomaly scenarios and a generated
  video corpus with ordinal action labels support exploratory evaluation
  of this two-stage monitor. With a fixed $T-2.5$\,s window, the proposed framework obtains $0.830$ accuracy and $0.826$ balanced
  accuracy versus $0.745$ and $0.732$ for a single-stage video-only
  baseline, increasing anomaly recall from $0.536$ to $0.773$ while reducing
  specificity from $0.928$ to $0.880$. Stage-2 comparison is nearly
  deterministic given the expectation and motion class while atypical-hazard expectation generation remains as a future work.

  \begingroup
  \let\originalthebibliography\thebibliography
  \renewcommand{\thebibliography}[1]{%
    \originalthebibliography{#1}%
    \interlinepenalty=10000\relax
  }

  \endgroup
\end{document}